\documentclass[letterpaper]{article} 
\usepackage{aaai2026}  
\usepackage{times}  
\usepackage{helvet}  
\usepackage{courier}  
\usepackage[hyphens]{url}  
\usepackage{graphicx} 
\usepackage{natbib}  
\usepackage{caption} 
\usepackage{algorithm}
\usepackage{algorithmic}

\usepackage{newfloat}
\usepackage{listings}
\DeclareCaptionStyle{ruled}{labelfont=normalfont,labelsep=colon,strut=off} 
\floatstyle{ruled}
\newfloat{listing}{tb}{lst}{}
\floatname{listing}{Listing}
\usepackage{tikz}
\usetikzlibrary{arrows.meta, positioning, patterns} 
\usepackage{booktabs}
\usepackage{multirow}
\usepackage{subcaption}

\usepackage{amsmath, amssymb, amsfonts, amsthm}

\title{CertMask: Certifiable Defense Against Adversarial Patches\\via Theoretically Optimal Mask Coverage}
\author{
Xuntao Lyu\textsuperscript{\rm 1}\equalcontrib,
Ching-Chi Lin\textsuperscript{\rm 2}\equalcontrib,
Abdullah Al Arafat\textsuperscript{\rm 1, 3},
Georg von der Br\"uggen\textsuperscript{\rm 2},\\
Jian-Jia Chen\textsuperscript{\rm 2, 4},
Zhishan Guo\textsuperscript{\rm 1, 2}
}
\affiliations{
\textsuperscript{\rm 1}North Carolina State University;
\textsuperscript{\rm 3}Florida International University;\\
\textsuperscript{\rm 2}Technische Universit\"at Dortmund;
\textsuperscript{\rm 4}Lamarr Institute for AI and ML\\
\{xlyu5, zguo32\}@ncsu.edu,\;
\{chingchi.lin, georg.von-der-brueggen, jian-jia.chen\}@cs.tu-dortmund.de,\;
aarafat@fiu.edu
}

\usepackage{bibentry}

\begin{document}

\maketitle



\begin{abstract}
Adversarial patch attacks inject localized perturbations into images to mislead deep vision models. These attacks can be physically deployed, posing serious risks to real-world applications. In this paper, we propose CertMask, a certifiably robust defense that constructs a provably sufficient set of binary masks to neutralize patch effects with strong theoretical guarantees. While the state-of-the-art approach (PatchCleanser) requires two rounds of masking and incurs $O(n^2)$ inference cost, CertMask performs only a single round of masking with $O(n)$ time complexity, where $n$ is the cardinality of the mask set to cover an input image. Our proposed mask set is computed using a mathematically rigorous coverage strategy that ensures each possible patch location is covered at least $k$ times, providing both efficiency and robustness. We offer a theoretical analysis of the coverage condition and prove its sufficiency for certification. Experiments on ImageNet, ImageNette, and CIFAR-10 show that CertMask improves certified robust accuracy by up to +13.4\% over PatchCleanser, while maintaining clean accuracy nearly identical to the vanilla model.
\end{abstract}

\begin{figure*}[t]
\centering
\hspace*{-1.5cm}
\includegraphics[width=0.85\textwidth, trim=0 0.4in 0.35in 0, clip]{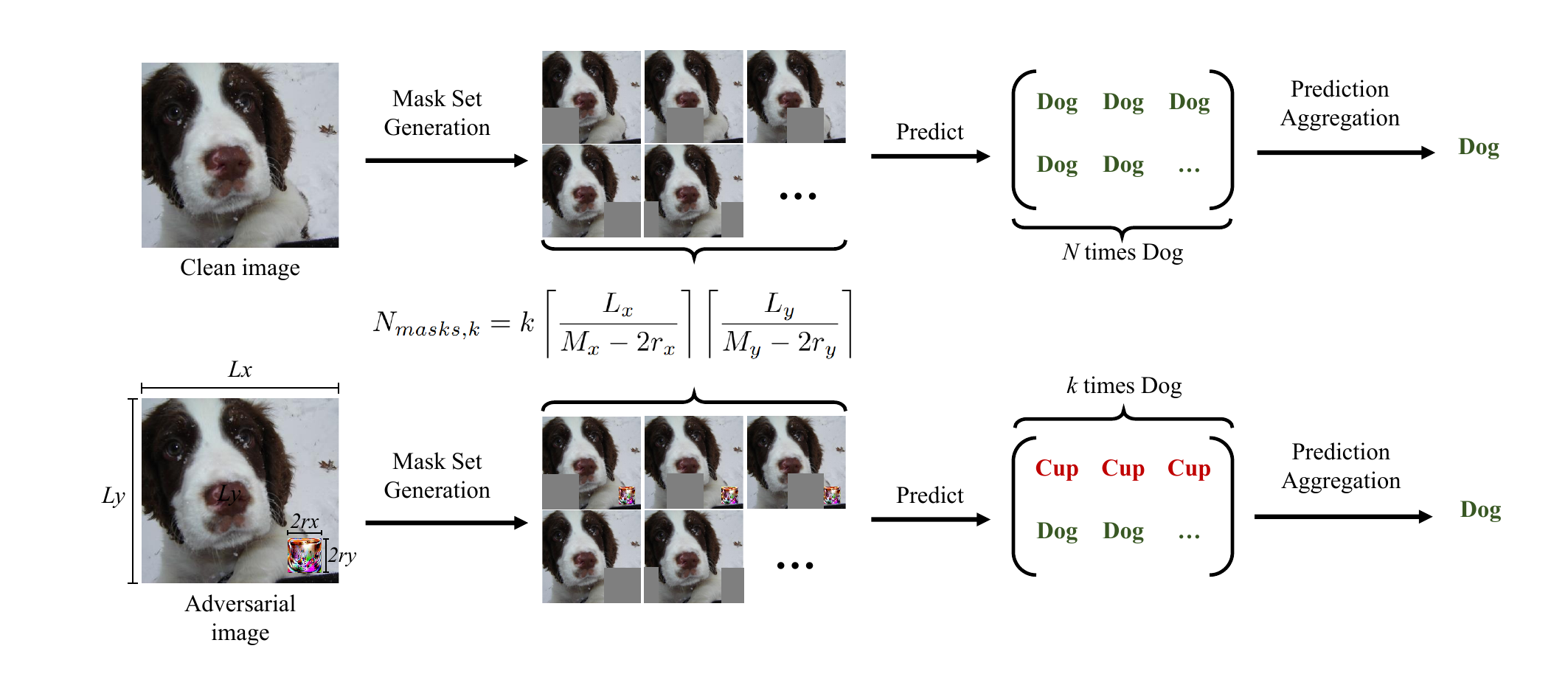}
\caption{\em Overview of the CertMask inference pipeline. Given an input image of size $L_x \times L_y$, we deterministically construct a set of $N_{\textit{masks,k}}$ binary masks, where each mask has spatial support $M_x \times M_y$ and is positioned such that every patch of size at most $2r_x \times 2r_y$ is guaranteed to be covered by exactly $k$ different masks. Each masked image is evaluated by the classifier, yielding a prediction. For aggregation, if all predictions agree, we output the unanimous result. If disagreement occurs and one class appears exactly $k$ times, that class is returned as the certified prediction. Otherwise, the majority class is selected to account for potential benign misclassifications.}
\label{fig:overview}
\end{figure*}

\section{Introduction}
Deep learning models have achieved strong performance in vision tasks such as robotics, autonomous driving, and surveillance. However, studies~\cite{xiao2018generating,yuan2022natural,shi2022decision,croce2022sparse} show that these models are vulnerable to adversarial attacks. Among them, adversarial patch attacks~\cite{brown2017adversarial,yang2020patchattack,karmon2018lavan} are particularly concerning due to their spatial locality. Unlike traditional perturbations that modify the entire input, patch attacks introduce small, confined regions that can significantly degrade performance. Moreover, they pose a practical threat because they can be physically instantiated and deployed in real-world scenes~\cite{yuan2024itpatch,xiao2021improving,wang2022defensive}.

Many existing defenses against adversarial patches require strong assumptions or are limited in practice. Some methods~\cite{levine2020randomized,lin2021certified,metzen2021efficient,xiang2021patchguard} rely on extracting intermediate activations from the model and enforcing architectural constraints, such as limiting the receptive field to reduce sensitivity to localized perturbations. These approaches typically assume access to the internal structure of the model and are restricted to specific architectures. Furthermore, constraining the receptive field can lead to a decline in clean accuracy, and the high computational cost of these methods often limits their applicability to low-resolution datasets or controlled offline settings. 

Certifiably robust defenses aim to overcome these limitations. They~\cite{xiang2022patchcleanser,saha2023revisiting,li2022vip} aim to provide provable guarantees that the model's prediction will remain unchanged under worst-case patch attacks. PatchCleanser~\cite{xiang2022patchcleanser}, one of the most effective examples, 
applies two rounds of masking to the input and aggregates predictions to certify robustness, is model-agnostic, and does not require retraining. However, PatchCleanser and similar defenses suffer from substantial computational overhead. In particular, PatchCleanser performs $O(n^2)$ forward passes  ($n$ is the number of masks to fully cover an input image) due to its pairwise masking strategy, making it impractical for high-resolution inputs or real-time deployment, where patch attacks can be executed instantly.

To address these limitations, we introduce CertMask, a novel certifiably robust defense that preserves provable robustness guarantees while achieving substantially improved computational efficiency. 
CertMask (illustrated in Figure~\ref{fig:overview}) deterministically allocates the provably sufficient number of masks required to ensure that any adversarial patch of bounded size is \emph{fully covered at least $k$ times} (a.k.a 
\textit{$k$-fold coverage}). 
By abstracting the patch coverage requirement into a discrete dot coverage problem, we derive the provably sufficient number of masks needed to ensure $k$-fold coverage for all potential patch locations. Both necessary and sufficient conditions for certified robustness are established.
Based on those conditions, we form two constructions 
with asymptotically optimal inference complexity of $O(n)$,  
a phenomenal improvement over prior $O(n^2)$ methods.


The remainder of the paper is organized as follows: Sec.~2 introduces the system model, threat model, and defines the problem. Sec. 3 derives the conditions for a patch to be fully covered, based on which CertMask is proposed and detailed in Sec. 4. Sec. 5 evaluates CertMask on multiple datasets and architectures, with ablation studies showing it achieves higher clean and certified robust accuracy than existing certifiable methods under standard settings. Sec. 6 concludes the work and points out future directions.

\section{Model and Problem}

This section introduces the system model, defines the threat model, and formally states the problem to be solved.

\subsection{System Model}



The \textbf{target domain} is a $2$-dimensional spatial region where an adversarial patch may appear, such as an input image. This domain is inherently \textbf{discrete}, composed of $L_x \times L_y$ pixels, 
forming a continuous rectangular area from $[0,L_x]$ and $[0,L_y]$ along the x- and y-axis, respectively. 

An \textbf{adversarial patch} is a localized region within the target domain, designed to mislead a deep learning model. We model this patch
as a solid, rectangular area with known dimensions but an \textbf{unknown location} within the target domain. Let the patch be centered at $(C_x, C_y)$ and have a total width of $2r_x$ and a total height of $2r_y$
(i.e., extending from $C_x-r_x$ to $C_x+r_x$ 
and from $C_y-r_y$ to $C_y+r_y$). 

A \textbf{mask} is a defensive mechanism applied to the target domain, intended to neutralize or obscure potential adversarial patches. In our model, a mask is a rectangular region of fixed dimensions, $M_x \times M_y$. 
The center of a mask 
can be positioned anywhere in the target domain 
(hence, its boundaries may partially extend beyond the defined target domain). When applied, a mask effectively leads to ignorance of the information within its boundaries, preventing it from influencing the deep learning model's output. 

\subsection{Threat Model}
We consider a test-time adversarial patch threat model. Let $f$ be a classifier and $\mathbf{x} \in \mathbf{R}^{L_x \times L_y \times C}$ (where  $L_x$ and $L_y$ denote the image width and height, and $C$ is the number of channels) be an input image with ground-truth label $y$. An adversary aims to construct an adversarial example $\mathbf{x}'$ such that $f(\mathbf{x}') \ne y$, by modifying a localized spatial region of $\mathbf{x}$.

The adversarial modification is constrained to a rectangular patch region $\Omega \subset [0, L_x] \times [0, L_y]$ with known size $2r_x \times 2r_y$, placed arbitrarily within the image. This constraint is represented by a binary mask $\mathbf{r} \in \{0,1\}^{L_x \times L_y \times C}$, where $\mathbf{r}_{i,j} = 1$ if $(i,j) \in \Omega$, and $0$ otherwise. The adversarial example is defined as:
\begin{equation}
    \mathbf{x}' = \mathbf{r} \odot \mathbf{z} + (1 - \mathbf{r}) \odot \mathbf{x},
\end{equation}
where $\mathbf{z} \in \mathbf{R}^{L_x \times L_y \times C}$ denotes arbitrary adversarial content, and $\odot$ is element-wise multiplication.

We make no assumptions about the attacker’s generation method or patch location. 
We only assume a known patch size, specifically that the adversarial region has dimensions $2r_x \times 2r_y$. This allows CertMask to defend against highly adaptive and unrestricted patch attacks without requiring access to patch placement or construction details.

\subsection{Adversarial Patch Covering (APC) Problem}\label{sec:problem_def}

We 
define the Adversarial Patch Covering \textbf{(APC)} problem as follows:
Consider a 2-dimensional 
spatial domain, representing an image or sensor input, which may contain a single adversarial patch of unknown location. The objective is to determine the minimum number of identical rectangular masks that must be strategically applied to this domain to guarantee that the patch is fully covered by at least $k$ of these masks. An adversarial patch is \textbf{fully covered} by a mask if and only if the entire area of the patch is strictly contained within the boundaries of the mask (so it can be neutralized). 

 
\paragraph{Definition 1 (Fully Covered):}\label{def:fully_covered}
A mask spanning from $[x, x+M_x]$ and $[y, y+M_y]$ fully covers an adversarial patch centered at $(C_x,C_y)$ with radii $r_x$, $r_y$ if: 
\begin{align}    
    x \leq C_x - r_x~&\mbox{and}~C_x + r_x \leq x + M_x,~\mbox{and} \nonumber \\
    y \leq C_y - r_y~&\mbox{and}~C_y + r_y \leq y + M_y \nonumber
\end{align}



For analytical tractability, we assume that all masks are of identical, rectangular dimensions, that the masks are sufficiently large to cover the adversarial patch, and that the adversarial patch itself is also rectangular.

This problem directly relates to practical challenges in ensuring the robustness of deep learning models against adversarial patch attacks. By finding the minimum number of masks needed to cover an unknown adversarial patch at least $k$ times, we are essentially determining an efficient masking strategy to neutralize its effect, thus preventing the attack from influencing the model's output without incurring excessive computational cost.

\section{Covering an Adversarial Patch}

We start by examining patch coverage in a 1-dimensional domain, then extend this analysis to 2-dimensional scenarios. We derive the conditions required for a patch to be fully covered by a mask---they are fundamental building blocks for developing effective solutions to the APC problem.

\subsection{APC in the 1-Dimensional Domain}

We begin by considering patch covering in a discrete \mbox{1-dimensional} domain. Assume an adversarial patch with radius $r$ centered at $C$, spanning the interval $[C-r,C+r]$. A mask of size $M > 2r$ is used for covering. To fully cover this patch, the mask's left endpoint, $x$, must satisfy $x \leq C-r$, while its right endpoint, $x+M$, must satisfy $x+M \geq C+r$.

As shown below, this patch-covering problem can be simplified into a dot-covering problem, where the full coverage of a patch is determined by the position of its center $C$.

\paragraph{Theorem 1.}\label{sec:Theorem_1}
    An adversarial patch centered at $C$ is fully covered by a mask $[x, x+M]$ if and only if $x+r \leq C \leq x+M-r$.
\paragraph{Proof.}
Provided in the supplementary documents.

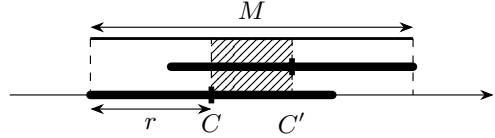
\begin{figure}
	\centering
    \resizebox{.8\linewidth}{!}{


\begin{tikzpicture}[yscale=0.7]
    \tikzset{
        every node/.style={font=\small}, 
        myarrow/.style={->, >=Stealth}, 
        dashedline/.style={dashed}, 
        boldline/.style={line width=3pt} 
    }

    \draw [myarrow] (0,0) -- (6,0);

    \draw [boldline, line cap=round] (1,0) -- (4,0); 
    \node at (2.5,-0.5) {\textbf{$C$}};    
    \node at (1.75,-0.5) {\textbf{$r$}};
    \draw [<->, >=Stealth] (1,-0.25) -- (2.5,-0.25);
    \draw [-, line width=2pt] (2.5,-0.15) -- (2.5,0.15);

    \draw [boldline, line cap=round] (2,0.5) -- (5,0.5); 
    \node at (3.5,-0.5) {\textbf{$C'$}};        
    \draw [-, line width=2pt] (3.5,0.65) -- (3.5,0.35);

    \draw [line width=1pt] (1,1) -- (5,1);
    \draw [dashedline] (1,1) -- (1,0);
    \draw [dashedline] (5,1) -- (5,0);
    \draw [<->, >=Stealth] (1,1.2) -- (5,1.2);    
    \node at (3,1.5) {\textbf{$M$}};

    \draw [dashed, pattern=north east lines] (2.5,0) rectangle (3.5,1);

\end{tikzpicture}

    }
    \caption{\textbf{1-D Effective Coverage Interval}. \em For a mask (size~$M$) and adversarial patch (radius $r$), the effective coverage interval (shaded) has a size of $M-2r$, representing the range where full patch coverage is guaranteed.}
    \label{fig:effective_coverage_1D}
\end{figure}

Based on Theorem 1, we can now define an \emph{effective coverage} interval within a mask. If the center $C$ of an adversarial patch falls within this interval, the patch is guaranteed to be fully covered by the mask, as shown in Figure~\ref{fig:effective_coverage_1D}.

\subsection{APC in the 2-Dimensional Domain}

\begin{figure}
	\centering
    \resizebox{.8\linewidth}{!}{


\begin{tikzpicture}[scale=0.9]    
    \tikzset{
        every node/.style={font=\small}, 
        dashedline/.style={dashed}, 
        boldline/.style={line width=3pt} 
    }

    \draw (0,0) rectangle (8, 5);
    \draw [<->, >=Stealth] (0,-0.5) -- (8,-0.5);
    \draw [<->, >=Stealth] (-0.5,0) -- (-0.5,5);
    \node at (4,-0.8) {\textbf{$M_x$}};
    \node at (-0.8,2.5) {\textbf{$M_y$}};    
    
    \draw [boldline](0,5) rectangle (2,3);
    \draw [<->, >=Stealth] (1,5.25) -- (2,5.25);
    \draw [<->, >=Stealth] (2.25,4) -- (2.25,5);
    \node at (1.5,5.5) {\textbf{$r_x$}};
    \node at (2.5,4.5) {\textbf{$r_y$}};    
    \node at (1,4.25) {\textbf{$(C_x,C_y)$}};    

    \draw [dashed, pattern=north east lines](1,4) rectangle (7,1);
    \draw [<->, >=Stealth] (7.25,1) -- (7.25,4);
    \draw [<->, >=Stealth] (1,0.75) -- (7,0.75);
    \node at (8,2.5) {\textbf{$M_y-2r_y$}};
    \node at (4,0.5) {\textbf{$M_x-2r_x$}};    
    
\end{tikzpicture}

    }
    \caption{\textbf{2-D Effective Coverage Area}. \em The shaded region illustrates the effective coverage area for a mask of size $M_x \times M_y$. For an adversarial patch with radii $r_x$ and $r_y$, this area has dimensions $(M_x-2r_x)\times(M_y-2r_y)$.}
    \label{fig:effective_coverage_2D}
\end{figure}
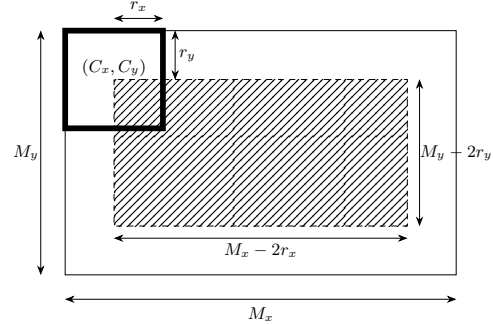

As established in Definition 1, an adversarial patch is \textbf{fully covered} by a mask if and only if its entire area is strictly contained within the mask's boundaries. This requires that the conditions for the patch's extent relative to the mask's boundaries are met independently along both the x- and y-axes in a \mbox{2-dimensional} domain.


Similar to the 1-D case, 2-D APC can be simplified by focusing on the center $(C_x, C_y)$ of the adversarial patch. Since the coverage in each dimension is independent, by 
Theorem 1, 
$x+r_x \leq C_x \leq x+M_x-r_x$ and $y+r_y \leq C_y \leq y+M_y-r_y$ are conditions for 2-D coverage.
These naturally lead to a rectangular \emph{effective coverage} area within a mask: if the center $(C_x, C_y)$ of the adversarial patch falls within this area, the entire patch is 
fully covered by the mask 
(see Figure~\ref{fig:effective_coverage_2D}).


\section{Methodology: CertMask}

This section presents our methodology for addressing the \emph{APC Problem}, which aims to minimize the number of masks applied while ensuring that any adversarial patch is fully covered by at least $k$ distinct masks, regardless of its unknown location. We first analyze the basic case of single coverage ($k=1$), and then extend our formulation to the general $k$-coverage setting.

\subsection{Single Cover}

\subsubsection{Single Cover in the 1-D domain.}

Theorem 1 shows that an adversarial patch of radius $r$ is fully covered by a mask of size $M$ if and only if the patch's center $C$ falls within the mask's effective coverage interval $[x+r,x+M-r]$. This effective coverage interval has a length of $M-2r$. Based on this crucial observation, we derive the following optimal pavement strategy for a \mbox{1-dimensional} domain $[0,L]$.

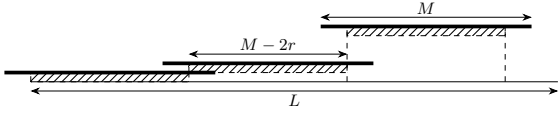
\begin{figure}
    \centering
    \resizebox{.9\linewidth}{!}{


\begin{tikzpicture}[yscale=0.7]
    \tikzset{
        every node/.style={font=\small}, 
        myarrow/.style={->, >=Stealth}, 
        dashedline/.style={dashed}, 
        boldline/.style={line width=3pt} 
    }

    \draw [-] (0,0) -- (10,0);
    \draw [<->, >=Stealth] (0,-0.25) -- (10,-0.25);
    \node at (5,-0.5) {\textbf{$L$}};

    \draw [line width=2pt] (-0.5,0.25) -- (3.5,0.25);
    \draw [dashed, pattern=north east lines] (0,0) rectangle (3,0.25);

    \draw [line width=2pt] (2.5,0.5) -- (6.5,0.5);
    \draw [dashed, pattern=north east lines] (3,0.25) rectangle (6,0.5);    
    \draw [dashedline] (6,0) -- (6,1.5);
    \draw [<->, >=Stealth] (3,0.75) -- (6,0.75);
    \node at (4.5,1) {\textbf{$M-2r$}};
    
    \draw [line width=2pt] (5.5,1.5) -- (9.5,1.5);
    \draw [dashed, pattern=north east lines] (6,1.25) rectangle (9,1.5);
    \draw [dashedline] (9,0) -- (9,1.5);
    \draw [<->, >=Stealth] (5.5,1.75) -- (9.5,1.75);
    \node at (7.5,2) {\textbf{$M$}};

\end{tikzpicture}

    }
    \caption{\textbf{Pavement strategy for 1-D domain}. \em This figure illustrates the mask placement strategy where effective coverage intervals are arranged contiguously, ensuring complete and gap-free coverage of the domain.}
    \label{fig:pavement_single_1D}
\end{figure}

Our strategy involves tiling the target domain with these effective coverage intervals, placing them contiguously without any gaps in the effective coverage as shown in Figure~\ref{fig:pavement_single_1D}. 
While the effective coverage intervals do not overlap, the 
masks themselves will inherently overlap by a length of $2r$ at their boundaries due to the $r$-offset from the mask edge to the effective coverage edge. 

To fully cover the target domain $[0,L]$ with these effective coverage intervals, each of length $M-2r$, the minimum number of masks required is $\lceil\frac{L}{M - 2r}\rceil$. We now formally prove that this quantity represents the optimal (minimum) number of masks for the 1-D APC 
with $k=1$.

\paragraph{Theorem 2.}
    To ensure that every possible position of an adversarial patch with length $2r$ within a 1-D domain of length $L$ is fully covered by at least one mask of size $M$, the number of masks $N_{masks}$ must satisfy:    
    \begin{equation}
        N_{masks} \geq \left\lceil\frac{L}{M-2r}\right\rceil
    \end{equation}
\paragraph{Proof.}
    Provided in the supplementary documents.

\subsubsection{Single Cover in the 2-D domain.}\label{sec:single_2D}

\begin{figure}
    \centering
    \resizebox{.6\linewidth}{!}{


\begin{tikzpicture}[scale=1]
    \tikzset{
        every node/.style={font=\small}, 
        myarrow/.style={->, >=Stealth}, 
        dashedline/.style={dashed}, 
        boldline/.style={line width=3pt} 
    }

    \draw (0,0) -- (-2.5,5);
    \draw (0,0) -- (10,0);

    \filldraw[        
        pattern=north east lines,
        line width=1pt,      
        dashed
    ] (0,0) -- (3,0) -- (2,2) -- (-1,2) -- cycle;
    \draw[line width=2pt] (-0.25,-0.3) -- (3.5,-0.3) -- (2.25,2.25) -- (-1.5,2.25) -- cycle;

    \begin{scope}[shift={(-1,2)}]
    \filldraw[        
        pattern=north east lines,
        line width=1pt,      
        dashed
    ] (0,0) -- (3,0) -- (2,2) -- (-1,2) -- cycle;
    \draw[line width=2pt] (-0.25,-0.3) -- (3.5,-0.3) -- (2.25,2.25) -- (-1.5,2.25) -- cycle;
    \end{scope}

    \begin{scope}[shift={(3,0)}]
    \filldraw[        
        pattern=north east lines,
        line width=1pt,      
        dashed
    ] (0,0) -- (3,0) -- (2,2) -- (-1,2) -- cycle;
    \draw[line width=2pt] (-0.25,-0.3) -- (3.5,-0.3) -- (2.25,2.25) -- (-1.5,2.25) -- cycle;
    \end{scope}

    \begin{scope}[shift={(2,2)}]
    \filldraw[        
        pattern=north east lines,
        line width=1pt,      
        dashed
    ] (0,0) -- (3,0) -- (2,2) -- (-1,2) -- cycle;
    \draw[line width=2pt] (-0.25,-0.3) -- (3.5,-0.3) -- (2.25,2.25) -- (-1.5,2.25) -- cycle;
    \end{scope}

    \begin{scope}[shift={(6,3)}]
    \filldraw[        
        pattern=north east lines,
        line width=1pt,      
        dashed
    ] (0,0) -- (3,0) -- (2,2) -- (-1,2) -- cycle;
    \draw[line width=2pt] (-0.25,-0.3) -- (3.5,-0.3) -- (2.25,2.25) -- (-1.5,2.25) -- cycle;
    \draw [dashedline, line width=2pt] (0,0) -- (0,-3);
    \draw [dashedline, line width=2pt] (-1,2) -- (-1,-1);
    \end{scope}

\end{tikzpicture}

    }
    \caption{\textbf{Pavement strategy for 2-D domain}. \em This figure illustrates the mask placement strategy where effective coverage areas are arranged contiguously, ensuring complete and gap-free coverage of the 2-D domain.}
    \label{fig:pavement_single_2D}
\end{figure}
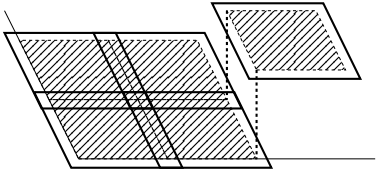

We extend the optimal pavement strategy to a 2-dimensional domain based on our 1-D findings. 
As established in our preliminary analysis, a patch is fully covered if and only if its center $(C_x, C_y)$ falls into the effective coverage area of one of the masks. For a 2-D mask, this effective coverage area is a rectangle of size $(M_x - 2r_x)(M_y - 2r_y)$.

An adversarial patch is covered by at least one mask on a 2-dimensional rectangular domain if and only if the union of the effective coverage areas of the deployed masks completely covers the entire $L_x \times L_y$ domain. Since each effective coverage area has dimensions $M_x - 2r_x$ by $M_y - 2r_y$, the total area to be covered is $L_xL_y$. Therefore, the theoretical minimum number of masks, $N_{masks}^{opt}$, must satisfy:

\begin{equation}
    N_{masks}^{opt} \geq \left\lceil \frac{L_xL_y}{(M_x - 2r_x)(M_y - 2r_y)} \right\rceil
\end{equation}

The pavement strategy for 2-D is a direct extension of the 1-D approach. We tile the 2-D domain by arranging masks such that their effective coverage areas are contiguously placed. Starting from the lower-left corner of the target domain, masks are applied row by row and column by column. Along the x-axis, masks are placed such that their effective coverage areas are adjacent, continuing until the total covered length exceeds $L_x$. The same principle applies along the y-axis. This strategy requires $\lceil\frac{L_x}{M_x - 2r_x}\rceil$ masks along the x-dimension and $\lceil\frac{L_y}{M_y - 2r_y}\rceil$ masks along the \mbox{y-dimension}. The total number of masks used by this strategy, $\lceil\frac{L_x}{M_x - 2r_x}\rceil\lceil\frac{L_y}{M_y - 2r_y}\rceil$, closely approximates the derived minimal bound.

\subsection{Multiple Cover}

Having established the mask deployment strategies for single-cover scenarios, we now extend our analysis to cases where an adversarial patch must be fully covered by more than one mask, specifically by at least $k$ masks ($k > 1$). Empirical evidence~\cite{mccoyd2020minority, xiang2022patchcleanser} suggests that increasing this coverage multiplicity $k$ can significantly enhance the robustness of deep learning models against adversarial patch attacks.


To achieve the desired $k$-fold coverage for an adversarial patch, we propose two distinct pavement strategies. 
\textbf{Replicated Tiling} is a straightforward extension 
of our single-cover solution and establishes a fundamental baseline. 
\textbf{Offset Tiling} provides an alternative and more intricate approach to mask placement, exploring more nuanced optimizations for practical deployment.


\subsubsection{Strategy 1: Replicated Tiling.}
From our analysis of 2-D single coverage, 
we established a $2$-approximation tiling strategy
that fully covers each adversarial patch at least once. 
A direct and intuitive approach to achieve $k$-fold coverage is to simply replicate this single-cover pavement plan $k$ times. This ensures that any adversarial patch, regardless of its location, will be fully covered by $k$ identical masks. 

We first establish a fundamental theoretical lower bound for the minimum number of masks required, $N_{masks, k}^{opt}$. Subsequently, we formally prove the approximation ratio of the \textbf{Replicated Tiling} strategy relative to this lower bound.

\paragraph{Theorem 3.}    
    To ensure that every possible position of a rectangular adversarial patch with size $2r_x \times 2r_y$ within a 2-D spatial domain of size $L_x \times L_y$ is fully covered by at least $k$ masks of size $M_x \times M_y$, the minimum number of masks $N_{masks,k}^{opt}$ must satisfy:
    \begin{equation}
        N_{masks,k}^{opt} \geq k\left\lceil \frac{L_xL_y}{(M_x - 2r_x)(M_y - 2r_y)} \right\rceil
    \label{eq:num_masks_rep}
    \end{equation}
\paragraph{Proof.}
Provided in the supplementary documents.

\paragraph{Theorem 4.}
    The \textbf{Replicated Tiling} strategy provides an approximation ratio of $2$.
\paragraph{Proof.}
Provided in the supplementary documents.

In practical implementation, instead of physically duplicating masks, this can correspond to applying the same masking pattern iteratively over $k$ different processing rounds or frames, or using $k$ independent masking layers.

\subsubsection{Strategy 2: Offset Tiling.}

We propose a compact and evenly distributed
mask placement strategy for $k$-fold coverage over a 2-D spatial domain termed Offset Tiling. Unlike Replicated Tiling, Offset Tiling interleaves multiple shifted mask grids within the same spatial domain to ensure that every adversarial patch is fully covered by at least $k$ different masks.



To achieve $k$-fold coverage, we decompose $k$ into two positive integers: $k = m \cdot n, ~\text{where, }m,n \in \mathbb{Z}^+$. Here, $m$ and $n$ are the desired number of overlapping masks along the horizontal and vertical axes. We then arrange the effective coverage regions of the masks on a uniform grid with horizontal and vertical strides:
\begin{align*}
    s_x = \frac{M_x - 2r_x}{m}; \quad
    s_y = \frac{M_y - 2r_y}{n}.
\end{align*}

We construct the tiling by placing masks such that the centers of their effective coverage areas are positioned at:
\(
(x_i, y_j) = (i \cdot s_x, j \cdot s_y), \quad \text{for integers } i, j \geq 0
\).
This guarantees that each point in the image will be guaranteed to be within the effective coverage area of at least $m$ horizontally placed masks and $n$ vertically placed masks, achieving the required $k$-fold coverage in total.

To ensure full coverage near the image boundaries, we adopt a wrap-around strategy: when a mask extends beyond the image border, its overflow wraps around to the opposite side (i.e., toroidal padding). This avoids coverage gaps at the edges and guarantees uniformity across the domain.

Total number of masks required under Offset Tiling is:
\begin{equation}
    N_{\text{masks},k} \geq \left\lceil \frac{m \cdot L_x}{M_x - 2r_x} \right\rceil \cdot \left\lceil \frac{n \cdot L_y}{M_y - 2r_y} \right\rceil
    \label{eq:num_masks_off}
\end{equation}

This approach provides a compact and efficient method for $k$-fold certified coverage, distributing masks more evenly than simple replication.


\begin{table*}[t]
\centering
\caption{\em 
Clean accuracy and certified robust accuracy (\%) under $k=6$ mask coverage ($k = 3 \times 2$). 
We use fixed square masks with side length 16 for 0.4\% patch, 32 for 1\% patch, 48 for 2\% patch, and 56 for both 2.4\% and 3\% patch. 
Bold highlights the best result among CertMask (CM) and prior defenses.
}

\label{tab:certified}
\setlength{\tabcolsep}{1.5pt}
\begin{tabular}{lcccccccccccccccc}
\toprule
\multirow{2}{*}{Method} &
\multicolumn{6}{c}{ImageNette} &
\multicolumn{6}{c}{ImageNet} &
\multicolumn{4}{c}{CIFAR-10} \\
\cmidrule(lr){2-7} \cmidrule(lr){8-13} \cmidrule(lr){14-17}
& \multicolumn{2}{c}{1\%} & \multicolumn{2}{c}{2\%} & \multicolumn{2}{c}{3\%}
& \multicolumn{2}{c}{1\%} & \multicolumn{2}{c}{2\%} & \multicolumn{2}{c}{3\%}
& \multicolumn{2}{c}{0.4\%} & \multicolumn{2}{c}{2.4\%} \\
& clean & robust & clean & robust & clean & robust
& clean & robust & clean & robust & clean & robust
& clean & robust & clean & robust \\
\midrule
CM-ResNet & \textbf{99.8} & \textbf{99.2} & \textbf{99.7} & 98.9 & \textbf{99.7} & 98.6
& 81.9 & 72.2 & 81.7 & 69.8 & 81.7 & 68.5 & 97.9 & 94.6
& 98.0 & 92.4 \\
CM-ViT & 99.6 & 99.1 & 99.6 & \textbf{99.0} & 99.6 & \textbf{98.9} & \textbf{84.1} & \textbf{77.6} & \textbf{84.5} & \textbf{75.5} & \textbf{84.5} & \textbf{74.0}
& \textbf{99.0} & \textbf{97.7} & \textbf{98.8} & 
\textbf{96.4} \\
CM-MLP & 99.4 & 98.9 & 99.4 & 98.7 & 99.4 & 98.5
& 80.2 & 72.1 & 80.0 & 69.8 & 79.8 & 67.9 & 97.5 & 93.8 & 97.1 & 90.8 \\
\midrule
PC-ResNet & \textbf{99.6} & 96.4 & \textbf{99.6} & 94.4 & \textbf{99.5} & 93.5
& 81.7 & 58.4 & 81.6 & 53.0 & 81.4 & 50.0 & 98.0 & 88.5
& 97.8 & 78.8 \\
PC-ViT & \textbf{99.6} & \textbf{97.5} & \textbf{99.6} & \textbf{96.4} & \textbf{99.5} & \textbf{95.3}
& \textbf{84.1} & \textbf{66.4} & \textbf{83.9} & \textbf{62.1} & \textbf{83.8} & \textbf{59.0} & \textbf{99.0} & \textbf{94.3}
& \textbf{98.7} & \textbf{89.1} \\
PC-MLP & 99.4 & 96.8 & 99.3 & 95.1 & 99.4 & 94.6
& 79.6 & 58.4 & 79.4 & 53.8 & 79.3 & 50.7 & 97.4 & 86.1
& 97.0 & 78.0 \\
CBN & 94.9 & 74.6 & 94.9 & 69.9 & 94.9 & 45.9
& 49.5 & 13.4 & 49.5 & 7.1 & 49.5 & 3.1 & 84.2 & 44.2
& 84.2 & 9.3 \\
DS & 92.1 & 82.3 & 92.1 & 79.1 & 92.1 & 75.7
& 44.4 & 17.7 & 44.4 & 14.0 & 44.4 & 11.2 & 83.9 & 68.9
& 83.9 & 56.2 \\
PG-BN & 95.2 & 89.0 & 95.0 & 86.7 & 94.8 & 83.0
& 55.1 & 32.3 & 54.6 & 26.0 & 54.1 & 19.7 & 84.5 & 63.8
& 83.9 & 47.3 \\
PG-DS & 92.3 & 83.1 & 92.1 & 79.9 & 92.1 & 76.8
& 44.1 & 19.7 & 43.6 & 15.7 & 43.0 & 12.5 & 84.7 & 69.2
& 84.6 & 57.7 \\
\bottomrule
\end{tabular}
\end{table*}

\begin{table}[t]
\centering
\caption{\em Clean accuracy of vanilla models.}
\label{tab:clean_acc}
\setlength{\tabcolsep}{6pt}
\begin{tabular}{lccc}
\toprule
  & ImageNette & ImageNet & CIFAR-10 \\
\midrule
ResNet & 99.8\% & 82.3\% & 98.3\% \\
ViT & 99.8\% & 84.8\% & 99.0\% \\
MLP & 99.5\% & 80.2\% & 97.8\% \\
\bottomrule
\end{tabular}
\end{table}

\section{Experiments}
\subsection{Setup}
In this section, we describe the evaluation setup including the datasets, models, attack configurations, evaluation metrics, and baseline defenses used to assess the effectiveness and robustness of our method.

\textbf{Datasets and Models.}
Three image classification benchmarks are used: ImageNet~\cite{deng2009imagenet}, ImageNette~\cite{imagenette2020}, and CIFAR-10~\cite{krizhevsky2009learning}. 
We use three image classification architectures: ResNetV2-50~\cite{he2016deep}, ViT-B/16~\cite{dosovitskiy2020image}, and ResMLP-S24~\cite{touvron2022resmlp}. All models are pretrained on ImageNet and fine-tuned with Cutout augmentation as in PatchCleanserw~\cite{xiang2022patchcleanser}.
We resize images to $224\times224$ via bicubic interpolation.

\textbf{Patch Attacks.}
Based on patch sizes commonly adopted in prior works~\cite{chiang2020certified,xiang2022patchcleanser}, we evaluate certified robustness against square patches occupying 1\%, 2\%, and 3\% of input pixels for ImageNet and ImageNette, and 0.4\% and 2.4\% for \mbox{CIFAR-10}. 
Patches are allowed at arbitrary locations with unrestricted content within image bounds.

\textbf{Metrics.}
We report 
(i)~\textbf{clean accuracy}, 
which is the fraction of unperturbed test images classified correctly, and (ii)~\textbf{certified robust accuracy}, 
which is the fraction of images where our certification procedure verifies prediction invariance against any patch within the threat model.

\textbf{Baselines.}
We compare with the state-of-the-art method PatchCleanser (PC)~\cite{xiang2022patchcleanser} and other certified defenses, namely 
PatchGuard (PG)~\cite{xiang2021patchguard}, Distribution Smoothing (DS)~\cite{levine2020randomized}, and Clipped BagNet (CBN)~\cite{zhang2020clipped}, using their best reported configurations.

\subsection{Results}

\subsubsection{Clean and certified robust accuracy.}
Table~\ref{tab:certified} reports the clean accuracy and certified robust accuracy of CertMask (CM), compared against the baseline approaches. 
All methods are evaluated under a unified protocol, using the same models, datasets, and patch sizes. For each patch threat level, we adopt a square mask with side length 16, 32, 48, or 56 to correspond to 0.4\%, 1\%, 2\%, and \{2.4\%, 3\%\} pixel patches, respectively. A fixed $3 \times 2 = 6$-fold mask coverage is implemented using our Offset Tiling strategy. 
For fair comparison, we use the same attack settings across methods, and highlight in bold the best certified accuracy achieved by both CertMask and the baseline.

CertMask consistently achieves competitive or superior clean accuracy across all datasets and architectures. As shown in Table~\ref{tab:certified}, CM-ViT attains 99.6\% clean accuracy on ImageNette and 99.0\% on CIFAR-10, matching PC-ViT. These values are also close to the clean performance of the corresponding undefended models reported in Table~\ref{tab:clean_acc}, indicating that CertMask introduces minimal accuracy degradation. For example, CM-ViT maintains only a 0.2\% drop from the vanilla 99.8\% on ImageNette and exactly matches the baseline performance on CIFAR-10.

\subsubsection{Certified robustness with minimal occlusion.}
In terms of certified robust accuracy, CertMask achieves substantial improvements over all prior methods. On ImageNet with a 2\%-pixel adversarial patch, CM-ViT achieves 75.5\% certified accuracy, surpassing PC-ViT (62.1\%) by +13.4\% and PG-DS (15.7\%) by +59.8\%. Similarly, on CIFAR-10 under a 2.4\%-pixel patch, CM-ViT achieves 96.4\%, outperforming PC-ViT (89.1\%) by +7.3\%. Similar gains are observed for ResNet and MLP 
across all datasets and threat models, 
reflecting the high efficacy of our mask construction.

A key factor 
for these gains is CertMask's single-round deterministic tiling strategy. Unlike PatchCleanser, which applies two rounds of randomized masking and requires $O(n^2)$ forward passes, CertMask ensures exact $k$-fold patch coverage in just one deterministic pass, using only $O(n)$ masks. This significantly reduces the total occlusion area, preserving more semantic context for prediction and improving robustness against patch attacks. Furthermore, the deterministic design eliminates the randomness-induced certification failures observed in prior stochastic methods.

\subsubsection{Cross-architecture and cross-dataset generality.} CertMask demonstrates strong generalization across architectures and datasets. On ImageNette with a 2\%-pixel patch, CertMask improves ViT’s certified accuracy from 96.4\% (PC-ViT) to 99.0\%, and boosts ResNet’s robustness from 94.4\% to 98.9\%. On ImageNet, CM-ViT improves certified robustness by over 10\% absolute margin across all tested patch sizes. These improvements are consistent across CIFAR-10 as well, 
showing CertMask’s adaptability to both low- and high-resolution images under varying threat levels.

Overall, CertMask sets a new state-of-the-art for certified patch robustness. Its deterministic, geometry-aware mask deployment achieves strong robust accuracy with minimal clean accuracy loss, offering a practical and certifiable solution for patch-resilient vision systems.

\begin{table}[t]
\centering
\caption{\em Effect of Tiling Strategy on Performance.}
\label{tab:tiling_comparison}
\setlength{\tabcolsep}{3pt}
\renewcommand{\arraystretch}{1.1}
\small
\begin{tabular}{llcccccc}
\toprule
\multirow{2}{*}{Model} & \multirow{2}{*}{Tiling} &
\multicolumn{2}{c}{ImageNette} &
\multicolumn{2}{c}{ImageNet} &
\multicolumn{2}{c}{CIFAR-10} \\
\cmidrule(lr){3-4} \cmidrule(lr){5-6} \cmidrule(lr){7-8}
& & clean & robust & clean & robust & clean & robust \\
\midrule
\multirow{2}{*}{ResNet}
  & Replicated & 99.5 & \textbf{99.3} & 77.9 & \textbf{73.7} & 96.1 & \textbf{94.9} \\
  & Offset & \textbf{99.7} & 98.6 & \textbf{81.7} & 68.5 & \textbf{98.0} & 92.4 \\
\midrule
\multirow{2}{*}{ViT}
  & Replicated & 99.4 & \textbf{99.2} & 82.0 & \textbf{78.6} & 98.4 & \textbf{97.8} \\
  & Offset & \textbf{99.6} & 98.9 & \textbf{84.5} & 74.0 & \textbf{98.8} & 96.4 \\
\midrule
\multirow{2}{*}{MLP}
  & Replicated & 99.2 & \textbf{99.0} & 76.7 & \textbf{72.5} & 95.7 & \textbf{94.1} \\
  & Offset & \textbf{99.4} & 98.5 & \textbf{79.8} & 67.9 & \textbf{97.1} & 90.8 \\
\bottomrule
\end{tabular}
\end{table}

\begin{table}[t]
\centering
\caption{\em Clean and robust accuracy under varying $k$.}
\label{tab:k}
\renewcommand{\arraystretch}{1.2}
\setlength{\tabcolsep}{6pt}
\small
\begin{tabular}{lcccccc}
\toprule
$k$ & 1 & 2 & 3 & 4 & 5 & 6 \\
\midrule
Clean (\%) & 82.0 & 80.8 & 82.5 & 83.4 & 83.8 & 84.5 \\
Robust (\%) & 78.6 & 77.0 & 76.2 & 74.9 & 75.7 & 74.0 \\
\bottomrule
\end{tabular}
\end{table}

\subsection{Detailed Parameter Analysis}

In this subsection, we systematically analyze how core design choices—tiling strategy, coverage multiplicity ($k$), mask size, and patch size—affect the clean and certified robust accuracy of CertMask.

\noindent{\bf Analysis of the performance of two Tiling Strategies.}
Table~\ref{tab:tiling_comparison} compares Replicated Tiling and Offset Tiling under identical evaluation settings. All experiments use a 3\% patch size for ImageNet and ImageNette, and 2.4\% for CIFAR-10, with a fixed mask size of 56. Both strategies ensure a $k$-fold coverage of $k = 6$. Replicated Tiling applies repeated mask grids at the same positions, while Offset Tiling shifts the masks to achieve complementary coverage.

Replicated Tiling consistently yields higher certified robust accuracy across all models and datasets. On ImageNet, it reaches 73.7\% with ResNet, outperforming Offset Tiling's 68.5\%; similar improvements are seen for ViT (78.6\% vs.\ 74.0\%) and MLP (72.5\% vs.\ 67.9\%). This robustness gain arises because the fixed mask layout avoids excessive occlusion of critical regions, while Offset Tiling’s shifted masks are more likely to disrupt semantically important areas.

Offset Tiling, however, achieves slightly better clean accuracy in some cases, such as ResNet on ImageNette (99.7\% vs.\ 99.5\%) and MLP on CIFAR-10 (98.0\% vs.\ 95.7\%), due to its more uniform occlusion pattern, which better preserves clean input information.
\begin{figure*}[t]
\centering
\begin{minipage}[t]{0.32\linewidth}
    \centering
    \includegraphics[width=\linewidth]{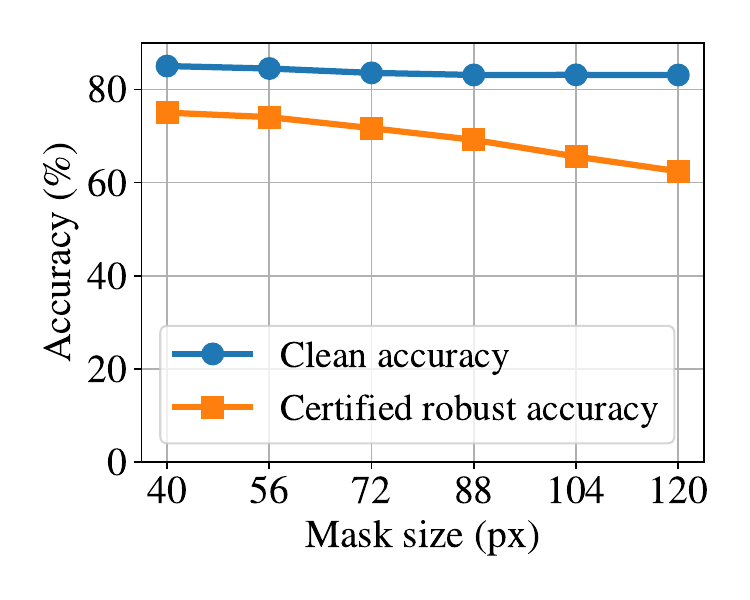}   
    \caption{\em The effect of mask size on defense performance.}
    \label{fig:effect_mask}
\end{minipage}
\hfill
\begin{minipage}[t]{0.32\linewidth}
    \centering
    \includegraphics[width=\linewidth]{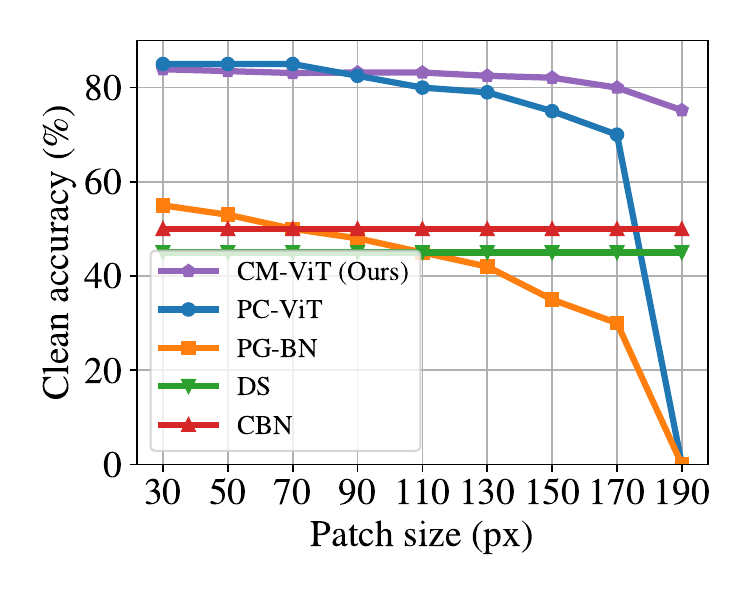} 
    \caption{\em Clean accuracy under different patch sizes.}
    \label{fig:clean_accuracy}
\end{minipage}
\hfill
\begin{minipage}[t]{0.32\linewidth}
    \centering
    \includegraphics[width=\linewidth]{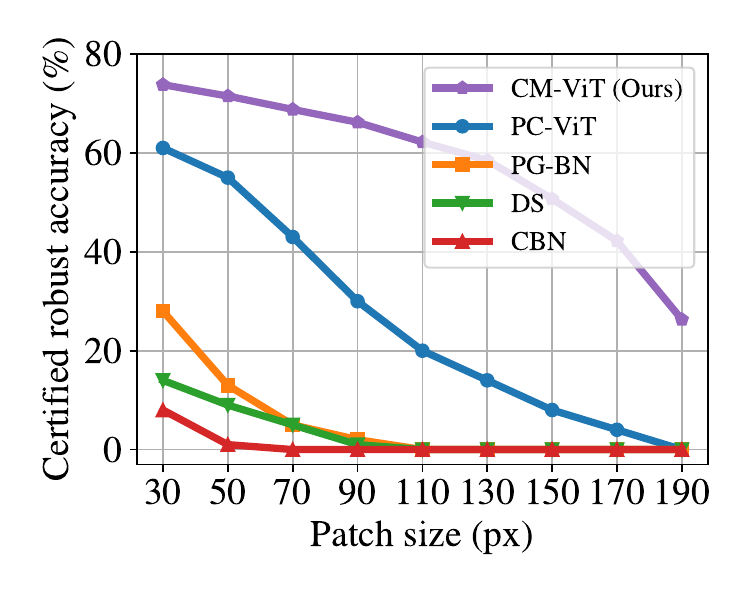}
    \caption{\em Certified robust accuracy under different patch sizes.}
    \label{fig:robust_accuracy}
\end{minipage}
\end{figure*}

\noindent{\bf Impact of $k$ on clean and certified robust accuracy.}
Table~\ref{tab:k} analyzes the effect of varying the mask coverage parameter $k$ on the clean and certified robust accuracy of CertMask using ViT on ImageNet. We fix the patch size to 3\% and mask size to 56, and sweep $k$ from 1 to 6 while maintaining the Offset tiling pattern.

As $k$ increases, the certified robust accuracy generally decreases, dropping from 78.6\% at $k=1$ to 74.0\% at $k=6$. This trend reflects the tradeoff introduced by heavier mask coverage: while larger $k$ values strengthen theoretical certification guarantees by covering each patch location more times, they also lead to increased occlusion, which can harm clean classification. Correspondingly, clean accuracy improves with larger $k$, increasing from 82.0\% at $k=1$ to 84.5\% at $k=6$, since higher $k$ reduces the number of masks per image, resulting in less aggressive masking per forward pass. This observation highlights that moderate values of $k$ (e.g., $k=3$ or $k=4$) may offer the best balance between certified robustness and clean performance.

\noindent{\bf Effect of Mask Size on Performance.}
Figure~\ref{fig:effect_mask} shows how varying the mask size affects clean and certified robust accuracy under a fixed mask coverage $k=6$ for ViT on ImageNet. As the mask size increases, certified robust accuracy drops steadily (e.g., from 76\% at 40px to 62\% at 120px), while clean accuracy remains relatively stable. This is because larger masks occlude more input area per mask, potentially hiding critical visual features needed for correct prediction. According to the theoretical bounds in Eq.~\ref{eq:num_masks_rep} and Eq.~\ref{eq:num_masks_off}, a larger mask size reduces the number of masks required to achieve $k$-fold coverage, thus lowering inference time. However, this comes at the cost of degraded robustness, suggesting a trade-off between computational efficiency and defense strength.

\noindent{\bf Performance under Varying Patch Sizes.}
Figures~\ref{fig:clean_accuracy} and~\ref{fig:robust_accuracy} illustrate the impact of increasing adversarial patch size on clean accuracy and certified robust accuracy, respectively. All evaluations are conducted on ViT with ImageNet under a fixed mask coverage of $k=6$.

As expected, the clean and certified accuracy of all methods degrade as the patch size increases. However, CertMask (CM-ViT) demonstrates significantly slower performance degradation compared to prior defenses. For example, when the patch size increases from 30 to 130 pixels, CM-ViT maintains clean accuracy above 80\% and certified accuracy around 66.8\%, whereas PatchCleanser (PC-ViT) rapidly declines to nearly 0\%. This robustness is attributed to CertMask’s single-round, image-agnostic tiling strategy: each image is processed with a fixed and carefully optimized mask set. In contrast, PatchCleanser generates one or two adaptive masks per image, which are more sensitive to the patch size—larger patches are more likely to escape randomized coverage, leading to faster performance degradation.

Although the accuracy of CertMask does decrease as the patch size grows, the drop is substantially slower, suggesting a higher tolerance to patch perturbation. This property is particularly valuable in realistic settings where the exact patch size is unknown or may vary, demonstrating that CertMask offers not only stronger robustness but also greater practical reliability under diverse threat conditions.

\section{Related Work}
Defenses against adversarial patch attacks can be broadly classified into empirical and certified approaches. Empirical methods~\cite{hayes2018visible,rao2020adversarial} evaluate robustness against specific attack strategies but lack formal guarantees and often fail under stronger or adaptive adversaries. In contrast, certified defenses~\cite{xiang2022patchcleanser,saha2023revisiting,li2022vip} aim to provide provable guarantees that model predictions remain invariant under any perturbation within a defined threat model. These methods are attack-agnostic and maintain their validity even under full adversarial knowledge, offering stronger and more reliable robustness at the expense of increased computational and design complexity.

\section{Conclusion}
In this work, we present CertMask, a mathematically grounded defense framework that certifies robustness against adversarial patch attacks via optimal mask coverage. By reducing the patch-covering problem to a geometric formulation, our approach constructs a provably sufficient set of masks that guarantees full patch coverage with controllable redundancy, achieving provable robustness with significantly improved computational efficiency. Compared to prior certified defenses, CertMask achieves superior clean and certified robust accuracy with linear certification cost, enabling scalable deployment to high-resolution vision tasks. As a promising future direction, our framework may be extended to spatiotemporal settings, enabling certified defenses for video models, as well as to scenarios where the adversarial patch size is unknown.

\section{Acknowledgment}
The research is partially supported by NSF grants CPS 2521121 and the Humboldt Fellowship. This work was supported in part by the German Federal Ministry of Education and Research (BMBF) in the Course of the 6GEM Research Hub under Grant 16KISK038.

\bibliography{aaai2026}

\section{A. Detailed Proofs of the Theorems}

\paragraph{Theorem 1.}\label{sec:Theorem_1}
    An adversarial patch centered at $C$ is fully covered by a mask $[x, x+M]$ if and only if $x+r \leq C \leq x+M-r$.
\paragraph{Proof.}
    \textbf{Sufficiency}: Assume that $C$ 
    falls in the interval \mbox{$[x+r, x+M-r]$}. The left endpoint of the patch is $C-r$. Since $C \geq x+r$, it follows that $C-r \geq x$. The right endpoint of the patch is $C+r$. Since $C \geq x+M-r$, it follows that $C+r \geq x+M$. Therefore, the interval $[C-r,C+r]$ is entirely contained within $[x,x+M]$, meaning the patch is fully covered.
  
    \textbf{Necessity}: Assume the adversarial patch, spanning \mbox{$[C-r, C+r]$}, is fully covered by the mask $[x, x+M]$. For the left endpoint of the patch to be covered:
    \begin{equation}
        C-r \geq x \Rightarrow{} C \geq x + r \nonumber
    \end{equation}    
    For the right endpoint of the patch to be covered:
    \begin{equation}
        C + r < x + M \Rightarrow{} C < x + M - r \nonumber
    \end{equation}
    Combining these two inequalities yields
    \begin{equation}
        x + r \leq C < x + M-r.
    \end{equation}
    
\paragraph{Theorem 2.}
    To ensure that every possible position of an adversarial patch with length $2r$ within a 1-D domain of length $L$ is fully covered by at least one mask of size $M$, the number of masks $N_{masks}$ must satisfy:    
    \begin{equation}
        N_{masks} \geq \left\lceil\frac{L}{M-2r}\right\rceil
    \end{equation}
\paragraph{Proof.}

    Assume, for contradiction, that we could achieve full coverage with $N_{masks}' < \lceil\frac{L}{M-2r}\rceil$ masks. The total combined length of the effective coverage provided by these $N_{masks}'$ masks would be $N_{masks}'(M-2r)$. By our assumption, $N_{masks}'(M-2r)<L$. This implies that there must exist at least one segment within the domain $[0,L]$ that is not covered by any mask's effective coverage. If the center $C$ of an adversarial patch were to fall within such an uncovered segment, that patch would not be fully covered by any of the $N_{masks}'$ masks, which contradicts our requirement. Therefore, $N_{masks}'$ cannot be smaller than $\lceil\frac{L}{M-2r}\rceil$. This completes the proof of optimality for the proposed pavement strategy.

\paragraph{Theorem 3.}    
    To ensure that every possible position of a rectangular adversarial patch with size $2r_x \times 2r_y$ within a 2-D spatial domain of size $L_x \times L_y$ is fully covered by at least $k$ masks of size $M_x \times M_y$, the minimum number of masks $N_{masks,k}^{opt}$ must satisfy:
    \begin{equation}
        N_{masks,k}^{opt} \geq k\left\lceil \frac{L_xL_y}{(M_x - 2r_x)(M_y - 2r_y)} \right\rceil
    \end{equation}
\paragraph{Proof.}
To ensure that every possible position of an adversarial patch within the domain is fully covered by at least $k$ masks, it implies that for every point $(C_x, C_y)$ in the domain, there must exist at least $k$ distinct masks whose effective coverage area includes $(C_x, C_y)$.

Let $A_{domain} = L_x L_y$ be the total area of the 2-D spatial domain. From our preceding analysis, we know that each mask provides an \textbf{effective coverage} area for the patch's center of $A_{eff}=(M_x-2r_x)(M_y-2r_y)$. This $A_{eff}$ represents the maximum area that a single mask can effectively cover for any patch center to guarantee full coverage of the patch itself.

Consider the total effective coverage capacity provided by $N_{masks,k}$ masks. This capacity is $N_{masks,k} A_{eff}$. For every point within the $L_x \times L_y$ domain to be covered by at least $k$ effective coverage areas, the sum of the areas effectively covered by all $N_{masks,k}$ masks, considering potential overlaps, must be at least $k$ times the total area of the domain. Mathematically, this can be expressed as:
\begin{align}
    \begin{split}
        &N_{masks,k} A_{eff} \geq k A_{domain}    \\
        \Rightarrow{}&N_{masks,k} (M_x-2r_x)(M_y-2r_y) \geq k L_x L_y  \nonumber  
    \end{split}    
\end{align}
Rearranging the inequality to solve for $N_{masks,k}^{opt}$ yields:
\begin{align}
    N_{masks,k}^{opt} \geq k \frac{L_x L_y}{(M_x-2r_x)(M_y-2r_y)}
\end{align}
Since $N_{masks,k}^{opt}$ is a positive integer, we have
\begin{align}
    N_{masks,k}^{opt} \geq k \left\lceil\frac{L_x L_y}{(M_x-2r_x)(M_y-2r_y)}\right\rceil.
\end{align}

This establishes the fundamental theoretical lower bound for k-fold coverage.

\paragraph{Theorem 4.}
    The \textbf{Replicated Tiling} strategy provides an approximation ratio of $2$
\paragraph{Proof.}
From the single-cover strategy in the 2-D domain, we know that the number of masks needed to cover the target domain once is $N_{masks,k=1} = \left\lceil\frac{L_x}{M_x - 2r_x}\right\rceil\left\lceil\frac{L_y}{M_y - 2r_y}\right\rceil$. By replicating this pavement for $k$ covers, the total number of masks used by Replicated Tiling is:
\begin{align}
    N_{masks, k} = k \left\lceil\frac{L_x}{M_x - 2r_x}\right\rceil\left\lceil\frac{L_y}{M_y - 2r_y}\right\rceil
    \label{eq:num_masks_rep}
\end{align}

From Theorem 3, the theoretical lower bound for the optimal number of masks $N_{masks,k}^{opt} = k\left\lceil \frac{L_xL_y}{(M_x - 2r_x)(M_y - 2r_y)} \right\rceil$.

To determine the approximation ratio $R = N_{masks, k} /  N_{masks, k}^{opt}$, we assume that $X = \frac{L_x}{M_x-2r_x}$ and $Y = \frac{L_y}{M_x-2r_y}$. Thus, we can simplifies $R$ into $\frac{\lceil X \rceil \lceil Y \rceil}{\lceil XY \rceil}$.

It is a known property in the field of geometric packing and covering problems that for any positive real numbers $X$ and $Y$, the ratio $\frac{\lceil X \rceil \lceil Y \rceil}{\lceil XY \rceil}$ is always less than or equal to $2$. This maximum ratio of $2$ occurs when $X$ and $Y$ are both infinitesimally greater than $1$ (e.g., $X=1+\epsilon$, $Y=1+\delta$, where $\epsilon,\delta \xrightarrow{}0^{+}$). In this specific scenario, $\lceil X \rceil = 2$ and $\lceil Y \rceil=2$, leading to $\lceil X \rceil \lceil Y \rceil=4$. Meanwhile, $XY=(1+\epsilon)(1+\delta)$, which for small $\epsilon$ and $\delta$ will be slightly greater than $1$. Thus, $\lceil XY \rceil=2$. The ratio in this worst-case scenario becomes $4/2=2$.

Therefore, $N_{masks,k} \leq 2 N_{masks,k}^{opt}$, establishing that Replicated Tiling has an approximation ratio of $2$.

\section{B. Detailed Experimental Setup}
\subsection{Details of Datasets}
\textbf{ImageNet} is a large-scale image classification dataset containing over 1.2 million training images across 1,000 object categories. Each image is a high-resolution natural image sourced from the internet, with substantial intra-class variation. In our experiments, we use the ILSVRC 2012 split and resize all images to $224 \times 224$ resolution. ImageNet serves as a representative benchmark for evaluating performance and robustness in complex, high-resolution scenarios.

\textbf{ImageNette} is a 10-class subset of ImageNet designed to be easier and faster to benchmark. It contains a small number of classes that are relatively well-separated semantically, reducing the label noise and ambiguity present in the full ImageNet dataset. We use the standard split and resize all images to $224 \times 224$. This dataset enables rapid iteration and analysis while maintaining visual diversity and realistic image complexity.

\textbf{CIFAR-10} is a widely used image classification dataset consisting of $60{,}000$ color images in $32 \times 32$ resolution, divided into 10 object classes. Each class contains $6{,}000$ images, with $50{,}000$ for training and $10{,}000$ for testing. Due to its small image size and standardized format, CIFAR-10 is commonly used for controlled experiments and ablation studies.

\subsection{Details of Model trainings}
We adopt a unified $224 \times 224$ input resolution across all datasets to ensure compatibility with standard vision backbones. For datasets with inherently high-resolution images (i.e., ImageNet and ImageNette), we resize and crop each image to fit this target size. For lower-resolution benchmarks CIFAR-10, we perform an upsampling step using bicubic interpolation without cropping.

All models are constructed using the \texttt{timm} framework, with ImageNet-pretrained weights used when applicable. For non-ImageNet datasets, we train models from scratch using a batch size of 64. Training is performed using SGD with a momentum of 0.9. The initial learning rate is 0.001 for ViT and ResMLP architectures, and 0.01 for ResNet. Each training run spans 10 epochs, and the learning rate is scheduled to decay by a factor of 10 every 5 epochs.

To enhance model resilience to occlusion-based perturbations, we employ Cutout augmentation as part of the default training configuration. In each training image, two square regions of size $128 \times 128$ are masked out at random locations. This augmentation is applied during training only, and is not used at test time.

\subsection{Details of Computing Infrastructure}
All experiments are performed on a compute server equipped with 48-core Intel Xeon Silver 4214 CPUs, 384 GB RAM, and 8 NVIDIA RTX A6000 GPUs (PCIe 4, 48 GB memory, compute capability 8.6). Our implementation is built on PyTorch, with all model architectures and pretrained weights managed through the timm library.

\section{C. Assumptions and Design Generality}
\paragraph{Mask Geometry and Patch Flexibility.}
The rectangular mask formulation is chosen for analytical clarity and computational simplicity. Importantly, this formulation does \emph{not} restrict the adversarial patch to a rectangular shape—it only constrains its maximum horizontal and vertical extents. Therefore, any patch whose projections along the $x$ and $y$ axes fall within the defined bounds remains certifiably covered by our masking strategy. The theoretical guarantees also extend to multiple localized patches, provided that they are not spatially dispersed beyond the coverage region.

\paragraph{Choice of $k$ and Practical Implications.}
The coverage replication factor $k$ is adjustable rather than fixed. As reported in Table~4, increasing $k$ generally improves clean accuracy while slightly reducing robust accuracy, accompanied by higher runtime and energy overhead. Since clean accuracy often better reflects real-world deployment outcomes, we recommend selecting the largest $k$ value that remains computationally feasible within system constraints.

\bigskip
\section{D. Extended Comparisons Analysis}
\label{app:comparisons}

This section provides extended quantitative comparisons and architectural analyses complementing the main results in Sec. 5. We evaluate CertMask across transformer- and convolution-based backbones under the standard 2\% patch perturbation setting.

\paragraph{Vision Transformer (ViT) Architectures.}
The most competitive existing transformer-based certified defense, MSGreedyCutout, reports $82.4\%$ clean and $64.9\%$ robust accuracy. Under identical conditions, CertMask achieves $84.5\%$ clean and $75.5\%$ robust accuracy, representing a substantial improvement of $+2.1\%$ in clean and $+10.6\%$ in robust accuracy. These results confirm that our theoretically grounded mask coverage strategy scales effectively to large transformer architectures while enhancing both certified and empirical robustness.

\paragraph{Convolutional Neural Network (CNN) Architectures.}
We further evaluate the generality of CertMask on convolutional architectures using ResNet backbones with different mask configurations. As shown in Fig.~6, the method achieves a clean accuracy of 81.7\% and a robust accuracy of 68.5\% for a mask size of 56, followed by 79.9\% and 61.4\% for a mask size of 88, and 79.5\% and 54.3\% for a mask size of 120. The observed decline in performance with increasing mask size is consistent with theoretical expectations, since larger masks reduce spatial resolution while introducing higher redundancy in coverage.

\end{document}